\documentclass[twoside,11pt]{article}

\usepackage{jmlr2e}
\usepackage{epsfig}
\usepackage{graphicx}
\usepackage{amsmath}
\usepackage{url} 
\usepackage{amssymb}
\usepackage{multirow}
\usepackage{algorithm}
\usepackage{graphicx}
\usepackage{fancyvrb}
\usepackage{color}
\usepackage{amsmath}
\usepackage{graphicx}
\usepackage{caption}
\usepackage{bbm}
\usepackage{algpseudocode}
\usepackage{float}
\usepackage{newverbs}
\usepackage{microtype}
\usepackage{xcolor}
\usepackage{breqn}
\usepackage{wrapfig}
\usepackage{soul}
\usepackage[colorinlistoftodos]{todonotes}
\usepackage[subrefformat=parens,labelformat=parens]{subfig}

\nipsfinalcopy % Uncomment for camera-ready version

\begin{document}

\title{Hierarchical Deep Reinforcement Learning: Integrating Temporal Abstraction and Intrinsic Motivation}

% \author{\name Marina Meil\u{a} \email mmp@stat.washington.edu \\
%       \addr Department of Statistics\\
%       University of Washington\\
%       Seattle, WA 98195-4322, USA
%       \AND
%       \name Michael I.\ Jordan \email jordan@cs.berkeley.edu \\
%       \addr Division of Computer Science and Department of Statistics\\
%       University of California\\
%       Berkeley, CA 94720-1776, USA}

% \author{\name Tejas D Kulkarni \email tejask@mit.edu \\
%       \addr Brain and Cognitive Science, \\
%       \addr Computer Science and Artificial Intelligence Laboratory, \\
%       Massachusetts Institute of Technology\\
%       Cambridge, MA 02139, USA}

% \author{
% David S.~Hippocampus\thanks{ Use footnote for providing further information
% about author (webpage, alternative address)---\emph{not} for acknowledging
% funding agencies.} \\
% Department of Computer Science\\
% Cranberry-Lemon University\\
% Pittsburgh, PA 15213 \\
% \texttt{hippo@cs.cranberry-lemon.edu} \\
% \And
% Coauthor \\
% Affiliation \\
% Address \\
% \texttt{email} \\
% \AND
% Coauthor \\
% Affiliation \\
% Address \\
% \texttt{email} \\
% \And
% Coauthor \\
% Affiliation \\
% Address \\
% \texttt{email} \\
% \And
% Coauthor \\
% Affiliation \\
% Address \\
% \texttt{email} \\
% (if needed)\\
% }

\author{Tejas D. Kulkarni\thanks{Authors contributed equally and listed alphabetically.} \\
		BCS, MIT \\
        \texttt{tejask@mit.edu}  \And
        Karthik R. Narasimhan$^*$ \\
        CSAIL, MIT \\
        \texttt{karthikn@mit.edu} \And
        Ardavan Saeedi \\ 
        CSAIL, MIT \\
        \texttt{ardavans@mit.edu} \And
        Joshua B. Tenenbaum \\
		BCS, MIT \\
        \texttt{jbt@mit.edu}
}

\maketitle

\begin{abstract}
% Learning goal-directed behavior in environments with sparse feedback is a major challenge for reinforcement learning algorithms. The primary difficulty arises due to insufficient exploration, resulting in the agent unable to learn robust value functions. Intrinsically motivated agents carry out behavior for its own sake rather than to directly solve problems. Such intrinsic behaviors could eventually help the agent solve goals posed by the environment. We present hierarchical-DQN (h-DQN), a framework to integrate hierarchical value functions, operating at different temporal scales, with deep reinforcement learning. A top-level value function learns a policy over intrinsic goals, and a lower-level function learns a policy over atomic actions to satisfy the given goals. h-DQN allows for flexible goal specifications, such as functions over entities and relations. This provides an efficient space for exploration in complicated environments. We demonstrate the strength of our approach on two problems with very sparse, delayed feedback: (1) a complex discrete MDP with stochastic transitions, and (2) the classic ATARI game called `Montezuma's Revenge'.
Learning goal-directed behavior in environments with sparse feedback is a major challenge for reinforcement learning algorithms. The primary difficulty arises due to insufficient exploration, resulting in an agent being unable to learn robust value functions. Intrinsically motivated agents can explore new behavior for its own sake rather than to directly solve problems. Such intrinsic behaviors could eventually help the agent solve tasks posed by the environment. We present hierarchical-DQN (h-DQN), a framework to integrate hierarchical value functions, operating at different temporal scales, with intrinsically motivated deep reinforcement learning. A top-level value function learns a policy over intrinsic goals, and a lower-level function learns a policy over atomic actions to satisfy the given goals. h-DQN allows for flexible goal specifications, such as functions over entities and relations. This provides an efficient space for exploration in complicated environments. We demonstrate the strength of our approach on two problems with very sparse, delayed feedback: (1) a complex discrete stochastic decision process, and (2) the classic ATARI game `Montezuma's Revenge'.
\end{abstract}

\section{Introduction}
Learning goal-directed behavior with sparse feedback from complex environments is a fundamental challenge for artificial intelligence. Learning in this setting requires the agent to represent knowledge at multiple levels of spatio-temporal abstractions and to explore the environment efficiently. Recently, non-linear function approximators coupled with reinforcement learning \cite{koutnik2014evolving,mnih2015human,silver2016mastering} have made it possible to learn abstractions over high-dimensional state spaces, but the task of exploration with sparse feedback still remains a major challenge. Existing methods like Boltzmann exploration and Thomson sampling~\cite{stadie2015incentivizing,osband2016deep} offer significant improvements over $\epsilon$-greedy, but are limited due to the underlying models functioning at the level of basic actions. In this work, we propose a framework that integrates deep reinforcement learning with hierarchical value functions (h-DQN), where the agent is motivated to solve intrinsic goals (via learning options) to aid exploration. These goals provide for efficient exploration and help mitigate the sparse feedback problem. Additionally, we observe that goals defined in the space of entities and relations can help significantly constrain the exploration space for data-efficient learning in complex environments.

Reinforcement learning (RL) formalizes control problems as finding a policy $\pi$ that maximizes expected future rewards \cite{sutton1998introduction}. Value functions $V(s)$ are central to RL, and they cache the utility of any state $s$ in achieving the agent's overall objective. Recently, value functions have also been generalized as $V(s,g)$ in order to represent the utility of state $s$ for achieving a given goal $g \in G$~\cite{sutton2011horde, schaul2015universal}. When the environment provides delayed rewards, we adopt a strategy to first learn ways to achieve intrinsically generated goals, and subsequently learn an optimal policy to chain them together. 
% Moreover, in high-dimensional problems, the value functions are typically represented by non-linear function approximators $V(s,g;\theta)$. 
Each of the value functions $V(s,g)$ can be used to generate a policy that terminates when the agent reaches the goal state $g$. A collection of these policies can be hierarchically arranged with temporal dynamics for learning or planning within the framework of semi-Markov decision processes~\cite{sutton1999between, szepesvari2014universal}. In high-dimensional problems, these value functions can be approximated by neural networks as $V(s,g;\theta)$. 

% Given raw pixel data, how could an agent learn to compose intrinsic goals in order to solve a given extrinsic goal?

We propose a framework with hierarchically organized deep reinforcement learning modules working at different time-scales. The model takes decisions over two levels of hierarchy -- (a) the top level module (\textit{meta-controller}) takes in the state and picks a new goal, (b) the lower-level module (\textit{controller}) uses both the state and the chosen goal to select actions either until the goal is reached or the episode is terminated. The \textit{meta-controller} then chooses another goal and steps (a-b) repeat. We train our model using stochastic gradient descent at different temporal scales to optimize expected future intrinsic (\textit{controller}) and extrinsic rewards (\textit{meta-controller}). We demonstrate the strength of our approach on problems with long-range delayed feedback: (1) a discrete stochastic decision process with a long chain of states before receiving optimal extrinsic rewards and (2) a classic ATARI game (`Montezuma's Revenge') with even longer-range delayed rewards where most existing state-of-art deep reinforcement learning approaches fail to learn policies in a data-efficient manner.
\section{Literature Review}

\subsection{Reinforcement Learning with Temporal Abstractions}
Learning and operating over different levels of temporal abstraction is a key challenge in tasks involving long-range planning. In the context of reinforcement learning~\cite{barto2003recent}, Sutton et al.\cite{sutton1999between} proposed the \emph{options} framework, which involves abstractions over the space of actions. At each step, the agent chooses either a one-step ``primitive'' action or a ``multi-step'' action policy (option). Each option defines a policy over actions (either primitive or other options) and can be terminated according to a stochastic function $\beta$. Thus, the traditional MDP setting can be extended to a semi-Markov decision process (SMDP) with the use of options. Recently, several methods have been proposed to learn options in real-time by using varying reward functions~\cite{szepesvari2014universal} or by composing existing options~\cite{sorg2010linear_options}. Value functions have also been generalized to consider goals along with states~\cite{schaul2015universal}. This universal value function $V(s,g;\theta)$ provides an universal option that approximately represents optimal behavior towards the goal $g$. Our work is inspired by these papers and builds upon them.  

There has also been a lot of work on option discovery in the tabular value function setting \cite{mcgovern2001automatic, csimcsek2005identifying, mannor2004dynamic, menache2002q}. In more recent work, Machado et al. \cite{machado2016learning} presented an option discovery algorithm where the agent is encouraged to explore regions that were previously out of reach. However, option discovery where non-linear state approximations are required is still an open problem. 

Other related work for hierarchical formulations include the model of Dayan and Hinton~\cite{dayan1993feudal} which consisted of ``managers'' taking decisions at various levels of granularity, percolating all the way down to atomic actions made by the agent. The MAXQ framework~\cite{dietterich2000hierarchical} built up on this work to decompose the value function of an MDP into combinations of value functions of smaller constituent MDPs, as did Guestrin et al.\cite{guestrin2003efficient} in their factored MDP formulation. Hernandez-Gardiol and Mahadevan~\cite{hernandez2001hierarchical} combined hierarchical RL with a variable length short-term memory of high-level decisions.
% provide another  solution to handle large state spaces with factored MDPs.

In our work, we propose a scheme for temporal abstraction that involves simultaneously learning options and a control policy to compose options in a deep reinforcement learning setting. Our approach does not use separate Q-functions for each option, but instead treats the option as part of the input, similar to  \cite{schaul2015universal}. This has two advantages: (1) there is shared learning between different options, and (2) the model is potentially scalable to a large number of options.

\subsection{Intrinsically motivated RL}
The nature and origin of `good' intrinsic reward functions is an open question in reinforcement learning. Singh et al.\cite{chentanez2004intrinsically} explored agents with intrinsic reward structures in order to learn generic options that can apply to a wide variety of tasks. Using a notion of ``salient events'' as sub-goals, the agent learns options to get to such events. In another paper, Singh et al.\cite{singh2010intrinsically} take an evolutionary perspective to optimize over the space of reward functions for the agent, leading to a notion of extrinsically and intrinsically motivated behavior. In the context of hierarchical RL, Goel and Huber~\cite{goel2003subgoal} discuss a framework for subgoal discovery using the structural aspects of a learned policy model. {\c{S}}im{\c{s}}ek et al.~\cite{csimcsek2005identifying} provide a graph partioning approach to subgoal identification.

Schmidhuber~\cite{schmidhuber2010formal} provides a coherent formulation of intrinsic motivation, which is measured by the improvements to a predictive world model made by the learning algorithm. Mohamed and Rezende~\cite{mohamed2015variational} have recently proposed a notion of intrinsically motivated learning within the framework of mutual information maximization. Frank et al.~\cite{frank2015curiosity} demonstrate the effectiveness of artificial curiosity using information gain maximization in a humanoid robot.

% For certain tasks (e.g. playing Atari games, building lego-blocks and other physical reasoning tasks), the space of goals can be narrowed down to the realm of entities and relations. This can enable the agent to learn optimal control policies for each goal, and higher level modules could then learn to compose these goals in order to maximize overall extrinsic reward. In arbitrary environments, the space of goals is unrestricted and unstructured, making it hard for such a scheme to observe non-zero reward values from the environment. The above mentioned task-agnostic measures should be combined with human-like goals for generality. 

\subsection{Object-based RL}
Object-based representations~\cite{diuk2008object,cobo2013object} that can exploit the underlying structure of a problem have been proposed to alleviate the \textit{curse of dimensionality} in RL. Diuk et al.\cite{diuk2008object} propose an  \textit{Object-Oriented MDP}, using a representation based on objects and their interactions. Defining each state as a set of value assignments to all possible relations between objects, they introduce an algorithm for solving deterministic object-oriented MDPs. Their representation is similar to that of Guestrin et al.\cite{guestrin2003generalizing}, who describe an object-based representation in the context of planning. In contrast to these approaches, our representation does not require explicit encoding for the relations between objects and can be used in stochastic domains. 

%\cite{diuk2008object} - object oriented RL

% Coupled with function approximation, we demonstrate the advantage of using a hierarchical model to learn in scenarios with significantly delayed, sparse rewards. 

\subsection{Deep Reinforcement Learning}
Recent advances in function approximation with deep neural networks have shown  promise in handling high-dimensional sensory input. Deep Q-Networks and its variants have been successfully applied to various domains including Atari games~\cite{mnih2015human} and Go~\cite{silver2016mastering}, but still perform poorly on environments with sparse, delayed reward signals. Strategies such as prioritized experience replay~\cite{schaul2015prioritized} and bootstrapping \cite{osband2016deep} have been proposed to alleviate the problem of learning from sparse rewards. These approaches yield significant improvements over prior work but struggle when the reward signal has a long delayed horizon. This is because the exploration strategy is not sufficient for the agent to obtain the required feedback.

\subsection{Cognitive Science and Neuroscience}

% In case of humans, goals could arise due to several reasons:

% could come from several sources in case of humans: (1) extrinsic behavior by the agent to maximize expected fitness function of the environment it inhabits (\cite{sutton1998introduction, singh2010intrinsically}), (2) goals communicated by other optimal agents (\cite{abbeel2004apprenticeship, baker2007goal, baker2008theory}), (3) intrinsic behavior by the agent to solve useful sub-problems via curiosity, play, etc.

% In this section, we mainly discuss (3). However, our approach is agnostic to the origin of sub-goals. 

The nature and origin of intrinsic goals in humans is a thorny issue but there are some notable insights from existing literature. There is converging evidence in developmental psychology that human infants, primates, children, and adults in diverse cultures base their core knowledge on certain cognitive systems including  -- entities, agents and their actions, numerical quantities, space, social-structures and intuitive theories~\cite{spelke2007core, lake2016building}. Even newborns and infants seem to represent the visual world in terms of coherent visual entities, centered around spatio-temporal principles of cohesion, continuity, and contact. They also seem to explicitly represent other agents, with the assumption that an agent's behavior is goal-directed and efficient. Infants can also discriminate relative sizes of objects, relative distances and higher order numerical relations such as the ratio of object sizes. During curiosity-driven activities, toddlers use this knowledge to generate intrinsic goals such as building physically stable block structures. In order to accomplish these goals, toddlers seem to construct sub-goals in the space of their core knowledge, such as -- putting a heavier entity \textit{on top of} (relation) a lighter entity in order to build tall blocks.

Knowledge of space can also be utilized to learn a hierarchical decomposition of spatial environments, where the bottlenecks between different spatial groupings correspond to sub-goals. This has been explored in neuroscience with the successor representation, which represents a value function in terms of the expected future state occupancy. Decomposition of the successor representation yields reasonable sub-goals for spatial navigation problems \cite{dayan1993improving, gershman2012successor,stachenfeld2014design}. Botvinick et al.\cite{botvinick2009hierarchically} have written a general overview of hierarchical reinforcement learning in the context of cognitive science and neuroscience.

\section{Model}
Consider a Markov decision process (MDP) represented by states $s \in \mathcal{S}$, actions $a \in \mathcal{A}$, and transition function $\mathcal{T}:(s, a) \rightarrow s'$. An agent operating in this framework receives a state $s$ from the external environment and can take an action $a$, which results in a new state $s'$. We define the extrinsic reward function as $\mathcal{F}:(s) \rightarrow \mathbb{R}$. The objective of the agent is to maximize this function over long periods of time. For example, this function can take the form of the agent's survival time or score in a game.

\begin{figure*}
    \centering
    \includegraphics[scale=0.5]{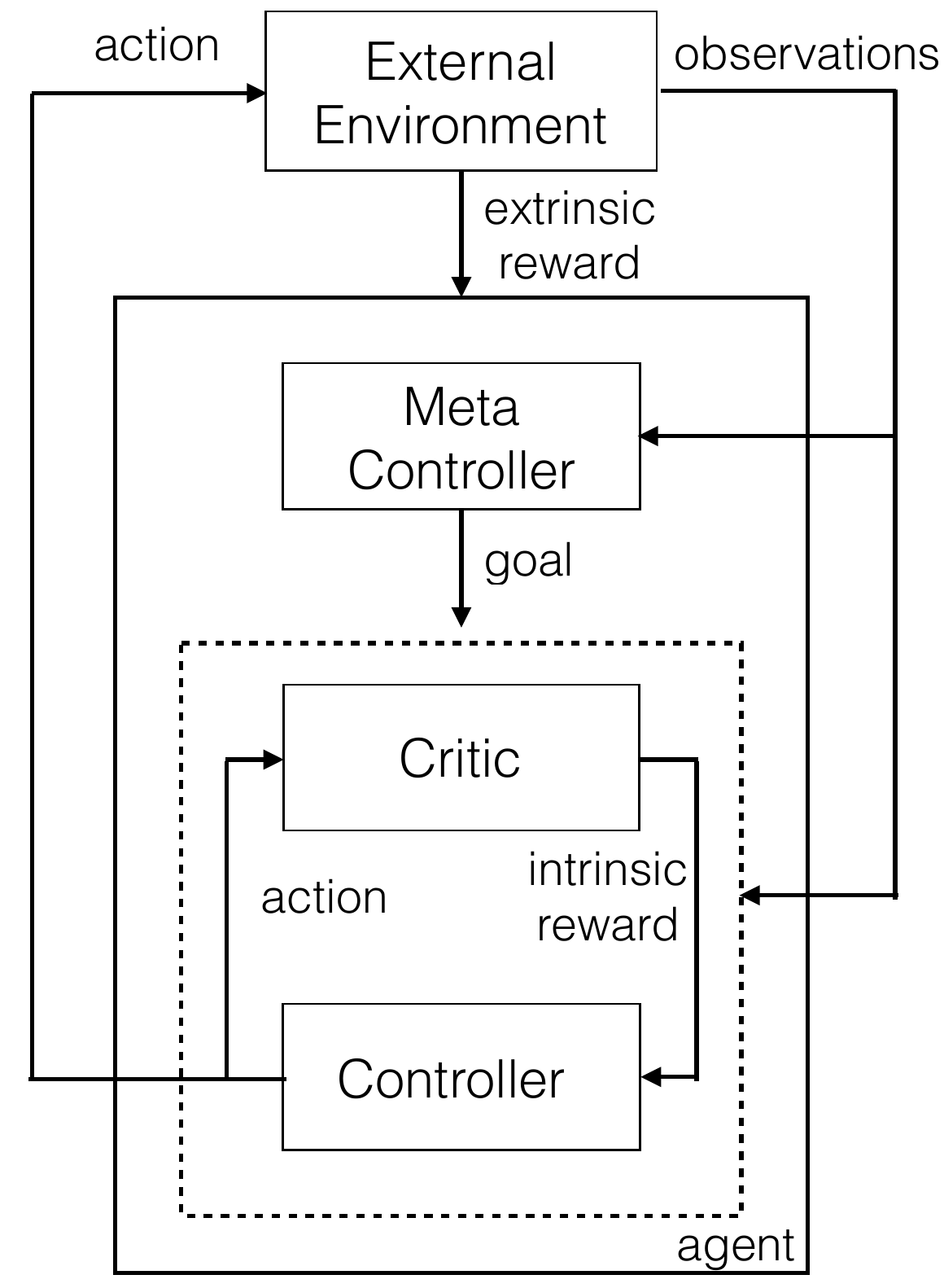} \\
    \vspace{10mm}
    \includegraphics[scale=0.55] {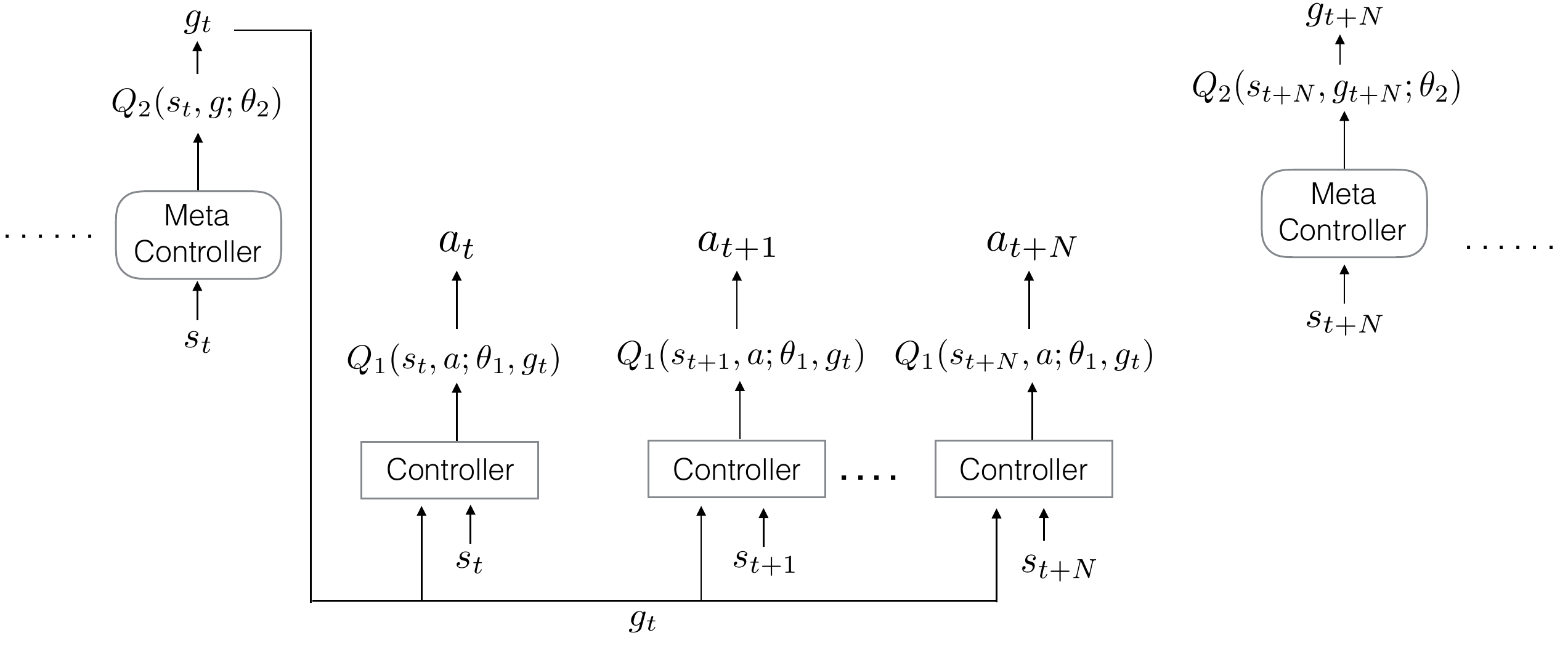}
      \caption{\textbf{Overview:} The agent produces actions and receives sensory observations. Separate deep-Q networks are used inside the \textit{meta-controller} and \textit{controller}. The meta-controller that looks at the raw states and produces a policy over goals by estimating the value function $Q_2(s_t,g_t;\theta_2)$ (by maximizing expected future extrinsic reward). The controller takes in states and the current goal, and produces a policy over actions by estimating the value function $Q_2(s_t,a_t;\theta_1, g_t)$ to solve the predicted goal (by maximizing expected future intrinsic reward).  The internal critic checks if goal is reached and provides an appropriate intrinsic reward to the controller. The controller terminates either when the episode ends or when $g$ is accomplished. The meta-controller then chooses a new $g$ and the process repeats. 
}
    \label{fig:overview}
\end{figure*}

\paragraph{Agents}
Effective exploration in MDPs is a significant challenge in learning good control policies. Methods such as $\epsilon$-greedy are useful for local exploration but fail to provide impetus for the agent to explore different areas of the state space. In order to tackle this, we utilize a notion of \emph{goals} $g \in \mathcal{G}$, which provide intrinsic motivation for the agent. The agent focuses on setting and achieving sequences of goals in order to maximize cumulative extrinsic reward.

We use the temporal abstraction of \emph{options}~\cite{sutton1999between} to define policies $\pi_g$ for each goal $g$. The agent learns these option policies simultaneously along with learning the optimal sequence of goals to follow. In order to learn each $\pi_g$, the agent also has a critic, which provides \emph{intrinsic rewards}, based on whether the agent is able to achieve its goals (see Figure~\ref{fig:overview}).

\paragraph{Temporal Abstractions}
As shown in Figure~\ref{fig:overview}, the agent uses a two-stage hierarchy consisting of a \emph{controller} and a \emph{meta-controller}. The meta-controller receives state $s_t$ and chooses a goal $g_t \in \mathcal{G}$, where $\mathcal{G}$ denotes the set of all possible current goals. The controller then selects an action $a_t$ using $s_t$ and $g_t$. The goal $g_t$ remains in place for the next few time steps either until it is achieved or a terminal state is reached. The internal critic is responsible for evaluating whether a goal has been reached and providing an appropriate reward $r_t(g)$ to the controller. The objective function for the controller is to maximize cumulative intrinsic reward: $R_t(g) = \sum_{t'=t}^{\infty} \gamma^{t'-t} r_{t'}(g)$. Similarly, the objective of the meta-controller is to optimize the cumulative extrinsic reward $F_t = \sum_{t'=t}^{\infty} \gamma^{t'-t} f_{t'}$, where $f_t$ are reward signals received from the environment. 

One can also view this setup as similar to optimizing over the space of optimal reward functions to maximize fitness~\cite{singh2009rewards}. In our case, the reward functions are dynamic and temporally dependent on the sequential history of goals. Figure~\ref{fig:overview} provides an illustration of the agent's use of the hierarchy over subsequent time steps.

% \begin{dmath}
% r(s, h_g) = \left\{
%                 \begin{array}{ll}
%                   r_g(s) \\
%                   \frac{1}{1+e^{-kx}}\\
%                   \frac{e^x-e^{-x}}{e^x+e^{-x}}
%                 \end{array}
%               \right.
% \end{dmath}

%  reward $r_{t'}^d$ denotes intrinsic reward $r_i$, otherwise it is supplied by the environment (extrinsic reward). We are interested in approximating the state-action value functions $Q^{*}_d(s,g_d)$ for all $d$: 

\subsection*{Deep Reinforcement Learning with Temporal Abstractions}
We use the Deep Q-Learning framework~\cite{mnih2015human} to learn policies for both the controller and the meta-controller. Specifically, the controller estimates the following Q-value function:
\begin{equation}
\begin{split}
Q^*_1(s,a;g) &= \max_{\pi_{ag}} \mathrm{E} \big[ \sum_{t'=t}^\infty \gamma^{t'-t} r_{t'}  \mid s_t=s, a_t=a, g_t=g, \pi_{ag} \big] \\
&= \max_{\pi_{ag}} \mathrm{E} \big[ r_t + \gamma~\text{max}_{a_{t+1}} Q^*_1(s_{t+1},a_{t+1}; g) \mid  s_t=s, a_t=a, g_t=g, \pi_{ag} \big ]
\end{split}
\end{equation}
where $g$ is the agent's goal in state $s$ and $\pi_{ag} = P(a|s,g)$ is the action policy.

Similarly, for the meta-controller, we have:
\begin{equation}
\begin{split}
% Q^*_2(s,g) &=\text{max}_{\pi_g} \mathrm{E} \big[ \sum_{t'=t}^\infty \gamma^{t'-t} f_{t'} \mid s_t=s, g_t=g, \pi_{g} \big] \\
Q^*_2(s,g) &=\text{max}_{\pi_g} \mathrm{E} \big[ \sum_{t'=t}^{t+N} f_{t'} + \gamma~\text{max}_{g'} Q_2^*(s_{t+N},g') \mid s_t=s, g_t=g, \pi_g \big]
\end{split}
\end{equation}
where $N$ denotes the number of time steps until the controller halts given the current goal, $g'$ is the agent's goal in state $s_{t+N}$, and $\pi_{g} = P(g|s)$ is the policy over goals. It is important to note that the transitions $(s_t, g_t, f_t, s_{t+N})$ generated by $Q_2$ run at a slower time-scale than the transitions $(s_t, a_t, g_t, r_t, s_{t+1})$ generated by $Q_1$. 

We can represent $Q^{*}(s,g) \approx Q(s, g;\theta)$ using a non-linear function approximator with parameters $\theta$, called a deep Q-network (DQN). Each $Q \in \{Q_1, Q_2\}$ can be trained by minimizing corresponding loss functions -- $L_{1}(\theta_{1})$ and $L_{2}(\theta_{2})$. We store experiences $(s_t, g_t, f_t, s_{t+N})$ for $Q_2$ and $(s_t, a_t, g_t, r_t, s_{t+1})$ for $Q_1$ in disjoint memory spaces $\mathcal{D}_1$ and $\mathcal{D}_2$ respectively. The loss function for $Q_1$ can then be stated as: 
\begin{equation}
\begin{split}
L_{1}(\theta_{1, i}) &= \mathrm{E}_{(s,a,g,r,s') \sim D_1} \big[ (y_{1,i} - Q_1(s,a;\theta_{1, i}, g))^2 \big],\\
\end{split}
\end{equation}
where $i$ denotes the training iteration number and $y_{1,i} =  r + \gamma~\text{max}_{a'} Q_1(s',a'; \theta_{1,i-1}, g)$. 

Following \cite{mnih2015human}, the parameters $\theta_{1, i-1}$ from the previous iteration are held fixed when optimising the loss function. The parameters $\theta_1$ can be optimized using the gradient: 
\begin{dmath*}
\centering
\nabla_{\theta_{1,i}} L_{1}(\theta_{1,i}) = {\mathrm{E}_{(s,a,r,s'\sim D_1)} \Bigg[ \Big(r + \gamma~\text{max}_{a'} Q_1(s',a'; \theta_{1,i-1}, g) - Q_1(s,a; \theta_{1,i}, g) \Big) 
\nabla_{\theta_{1,i}} Q_1(s,a; \theta_{1,i}, g) \Big)  \Bigg] }
\label{eq:updaterule}
\end{dmath*}
The loss function $L_2$ and its gradients can be derived using a similar procedure. 

\paragraph{Learning Algorithm}
We learn the parameters of h-DQN using stochastic gradient descent at different time scales -- experiences (or transitions) from the controller are collected at every time step but experiences from meta-controller are only collected when the controller terminates (i.e. when a goal is re-picked or the episode ends). Each new goal $g$ is drawn in an $\epsilon$-greedy fashion (Algorithms \ref{alg:training} \& \ref{alg:epsgreedy}) with the exploration probability $\epsilon_2$ annealed as learning proceeds (from a starting value of 1). 

In the controller, at every time step, an action is drawn with a goal using the  exploration probability $\epsilon_{1,g}$ which is dependent on the current empirical success rate of reaching $g$. The model parameters $(\theta_1,\theta_2)$ are  periodically updated by drawing experiences from replay memories $\mathcal{D}_1$ and $\mathcal{D}_2)$, respectively (see Algorithm~\ref{alg:updateparams}).

\begin{algorithm}[t]
\caption{Learning algorithm for h-DQN}
\label{alg:training}
\begin{algorithmic}[1]
\State Initialize experience replay memories $\{\mathcal{D}_1, \mathcal{D}_2\}$ and parameters $\{\theta_1,\theta_2\}$ for the controller and meta-controller respectively.
\State Initialize exploration probability $\epsilon_{1,g}=1$ for the controller for all goals $g$ and $\epsilon_2=1$ for the meta-controller.
\For {$ i = 1, num\_episodes $}
	\State Initialize game and get start state description $s$
	\State $g \leftarrow \textsc{epsGreedy}(s, \mathcal{G}, \epsilon_2, Q_2)$ 
% 	\State $prev\_s \leftarrow s$
    \While { $s$ is \textbf{not} terminal}
	\State $F \leftarrow 0$
    \State $s_0 \leftarrow s$
	\While { \textbf{not} ($s$ is terminal \textbf{or} goal $g$ reached)}       
	        \State $a \leftarrow \textsc{epsGreedy}(\{s,g\}, \mathcal{A}, \epsilon_{1,g}, Q_1)$
	        \State Execute $a$ and obtain next state $s'$ and extrinsic reward $f$ from environment
            \State Obtain intrinsic reward $r(s,a,s')$ from internal critic
            \State Store transition $(\{s,g\},a,r,\{s',g\})$ in $\mathcal{D}_1$
	        \State $\textsc{updateParams}(\mathcal{L}_1(\theta_{1,i}),\mathcal{D}_1$)
            \State $\textsc{updateParams}(\mathcal{L}_2(\theta_{2,i}),\mathcal{D}_2$)
	        \State $F \leftarrow F + f$
	        \State $s \leftarrow s'$
	    \EndWhile
	    \State Store transition $(s_0,g,F,s')$ in $\mathcal{D}_2$
        \If {$s$ is \textbf{not} terminal}
        	\State $g \leftarrow \textsc{epsGreedy}(s,\mathcal{G}, \epsilon_2,Q_2)$ 
        \EndIf
	\EndWhile
	\State Anneal $\epsilon_2$ and adaptively anneal $\epsilon_{1,g}$ using average success rate of reaching goal $g$.
\EndFor
\end{algorithmic}
\end{algorithm}

\begin{algorithm}[t]
\caption{: \textsc{epsGreedy}($x, \mathcal{B}, \epsilon, Q$)}
\label{alg:epsgreedy}
\begin{algorithmic}[1]
\If {random() $ < \epsilon$}
	\State \Return random element from set $\mathcal{B}$
\Else
    \State \Return $\text{argmax}_{m \in \mathcal{B}} \mathcal{} Q(x,m)$
\EndIf
\end{algorithmic}
\end{algorithm}

\begin{algorithm}[h!]
\caption{: \textsc{updateParams}($\mathcal{L},\mathcal{D})$}
\label{alg:updateparams}
\begin{algorithmic}[1]
% \State Store $(x,m,r,x')$ in $\mathcal{D}$
\State Randomly sample mini-batches from $\mathcal{D}$
\State Perform gradient descent on loss $\mathcal{L}(\theta)$ (cf. \eqref{eq:updaterule})
\end{algorithmic}
\end{algorithm}

\section{Experiments}
We perform experiments on two different domains involving delayed  rewards. The first is a discrete-state MDP with stochastic transitions, and the second is an ATARI 2600 game called `Montezuma's Revenge'.

\subsection {Discrete stochastic decision process}
% \begin{figure}[h]
\begin{wrapfigure}{R}{0.5\textwidth}
    \centering
    \includegraphics[width=0.48\textwidth]{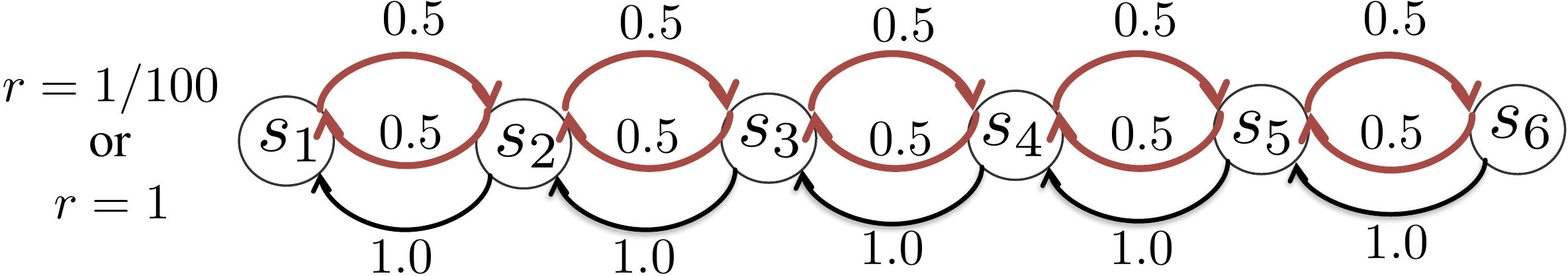}
    \caption{A stochastic decision process where the reward at the terminal state $s_1$ depends on whether $s_6$ is visited ($r=1$) or not ($r=1/100$).}
    \label{fig:syn_illus}
% \end{figure}
\vspace{-5mm}
\end{wrapfigure}

\paragraph{Game Setup} We consider a stochastic decision process where the extrinsic reward depends on the history of visited states in addition to the current state. We selected this task in order to demonstrate the importance of intrinsic motivation for exploration in such environments. 

There are 6 possible states and the agent always starts at $s_2$. The agent moves left deterministically when it chooses \emph{left} action; but the action \emph{right} only succeeds 50\% of the time, resulting in a left move otherwise. The terminal state is $s_1$ and the agent receives the reward of 1 when it first visits $s_6$ and then $s_1$. The reward for going to $s_1$ without visiting $s_6$ is 0.01. This is a modified version of the MDP in \cite{osband2016deep}, with the reward structure adding complexity to the task. The process is illustrated in Figure~\ref{fig:syn_illus}. 

We consider each state as a possible goal for exploration. This encourages the agent to visit state $s_6$ (whenever it is chosen as a goal) and hence, learn the optimal policy. For each goal, the agent receives a positive intrinsic reward if and only if it reaches the corresponding state.

\begin{figure}[h]
    \centering
    \includegraphics[scale=0.39]{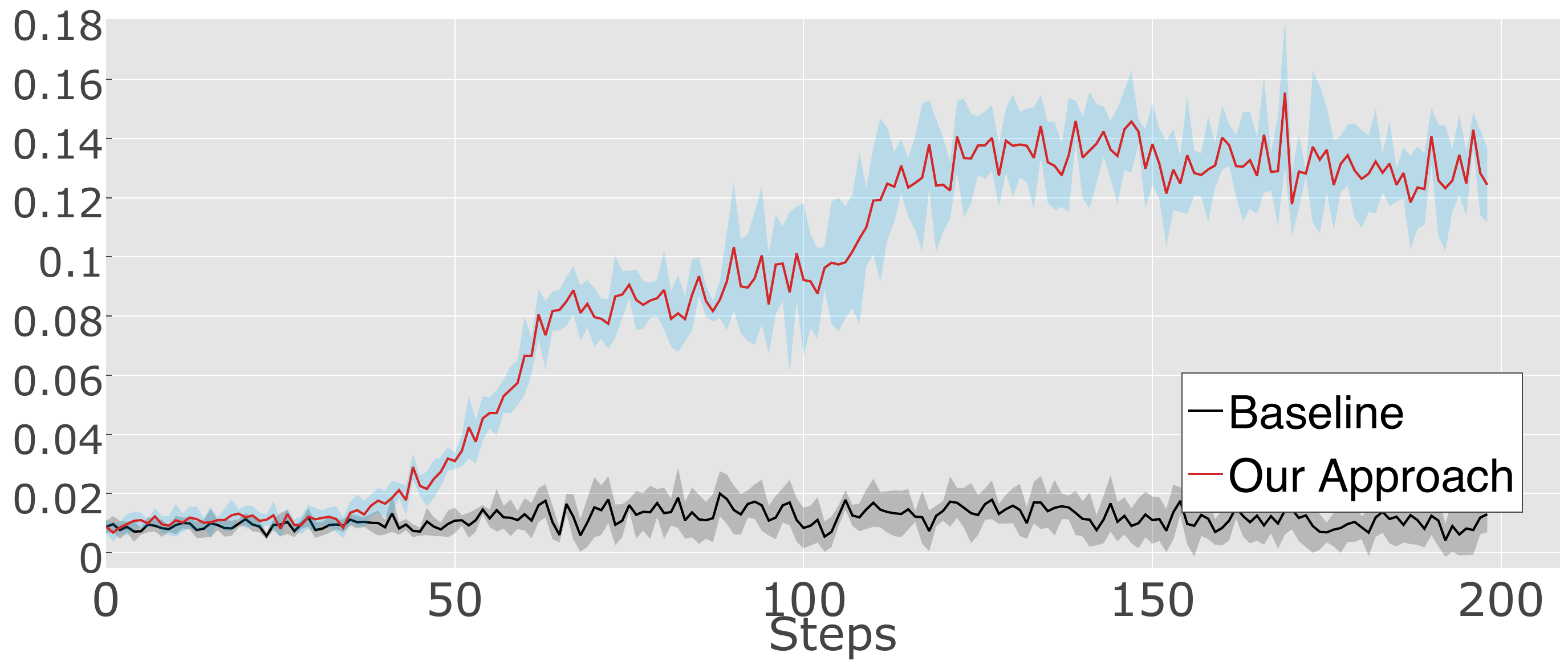}
    \caption{Average reward for 10 runs of our approach compared to Q-learning.}
    \label{fig:syn}
\end{figure}

\paragraph{Results} We compare the performance of our approach (without the deep neural networks) with Q-Learning as a baseline (without intrinsic rewards) in terms of the average extrinsic reward gained in an episode. In our experiments, all $\epsilon$ parameters are annealed from 1 to 0.1 over 50,000 steps. The learning rate is set to $0.00025$. Figure~\ref{fig:syn} plots the evolution of reward for both methods averaged over 10 different runs. As expected, we see that Q-Learning is unable to find the optimal policy even after 200 epochs, converging to a sub-optimal policy of reaching state $s_1$ directly to obtain a reward of 0.01. In contrast, our approach with hierarchical Q-estimators learns to choose goals $s_4$, $s_5$ or $s_6$, which statistically lead the agent to visit $s_6$ before going back to $s_1$.  Therefore, the agent obtains a significantly higher average reward of around 0.13.

~
\begin{figure}[h]
    \centering
    \includegraphics[scale=0.39]{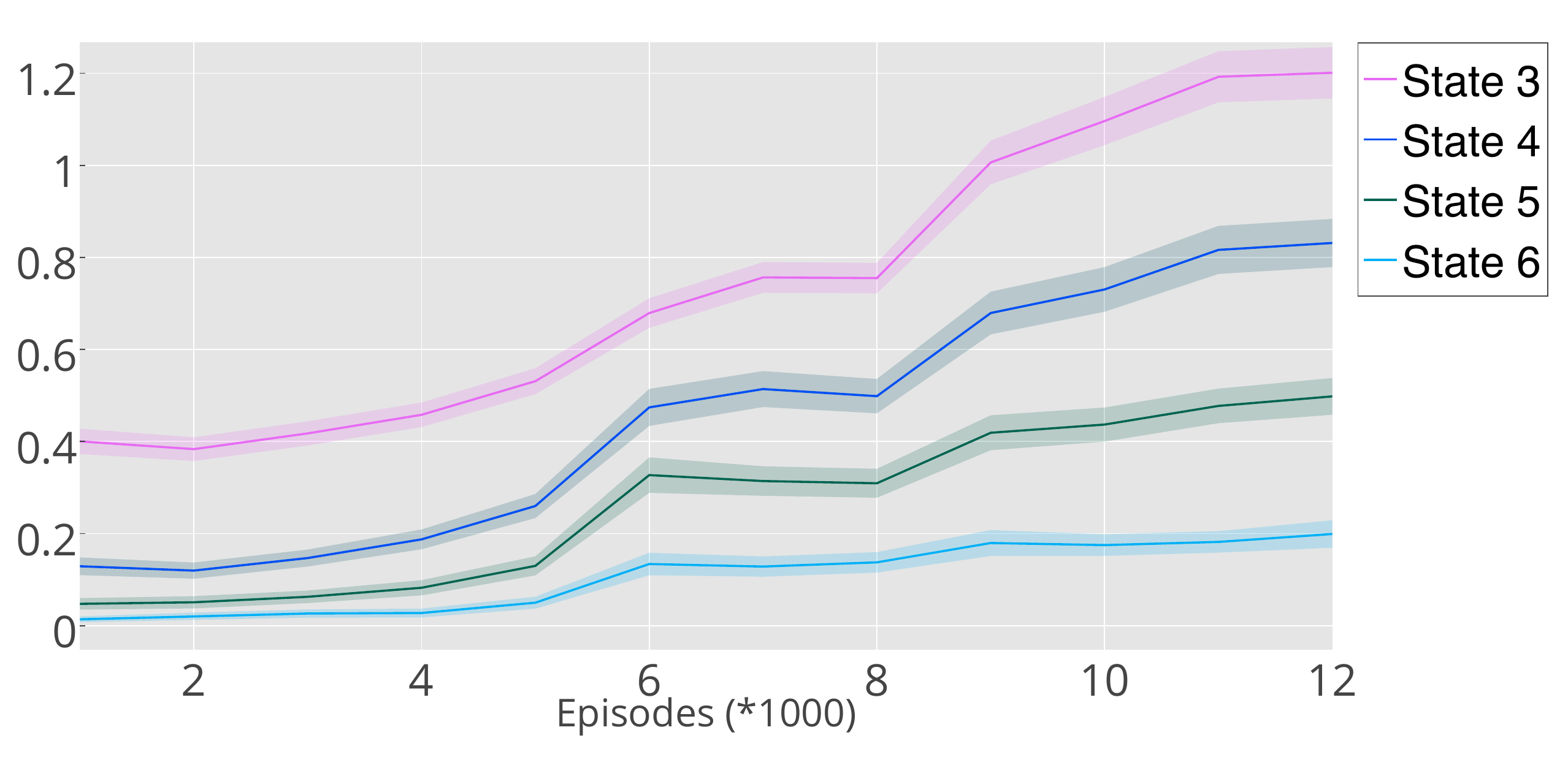}
    \caption{Number of visits (for states $s_3$ to $s_6$) averaged over 1000 episodes. The initial state is $s_2$ and the terminal state is $s_1$.}
    \label{fig:syn_goals}
\end{figure}

Figure~\ref{fig:syn_goals} illustrates that the number of visits to states $s_3, s_4, s_5, s_6$ increases with episodes of training. Each data point shows the average number of visits for each state over the last 1000 episodes. This indicates that our model is choosing goals in a way so that it reaches the critical state $s_6$ more often.     

% \textbf{TODO: add some subgoal picking stats to show that we learn to choose $s_6$.}

\subsection {ATARI game with delayed rewards}
\label{mzsec}
\begin{figure*}
\centering
    (a) \includegraphics[width=0.3\textwidth]{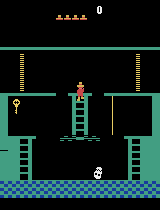} \hspace{1.3cm}
    (b) \includegraphics[width=0.5\textwidth]{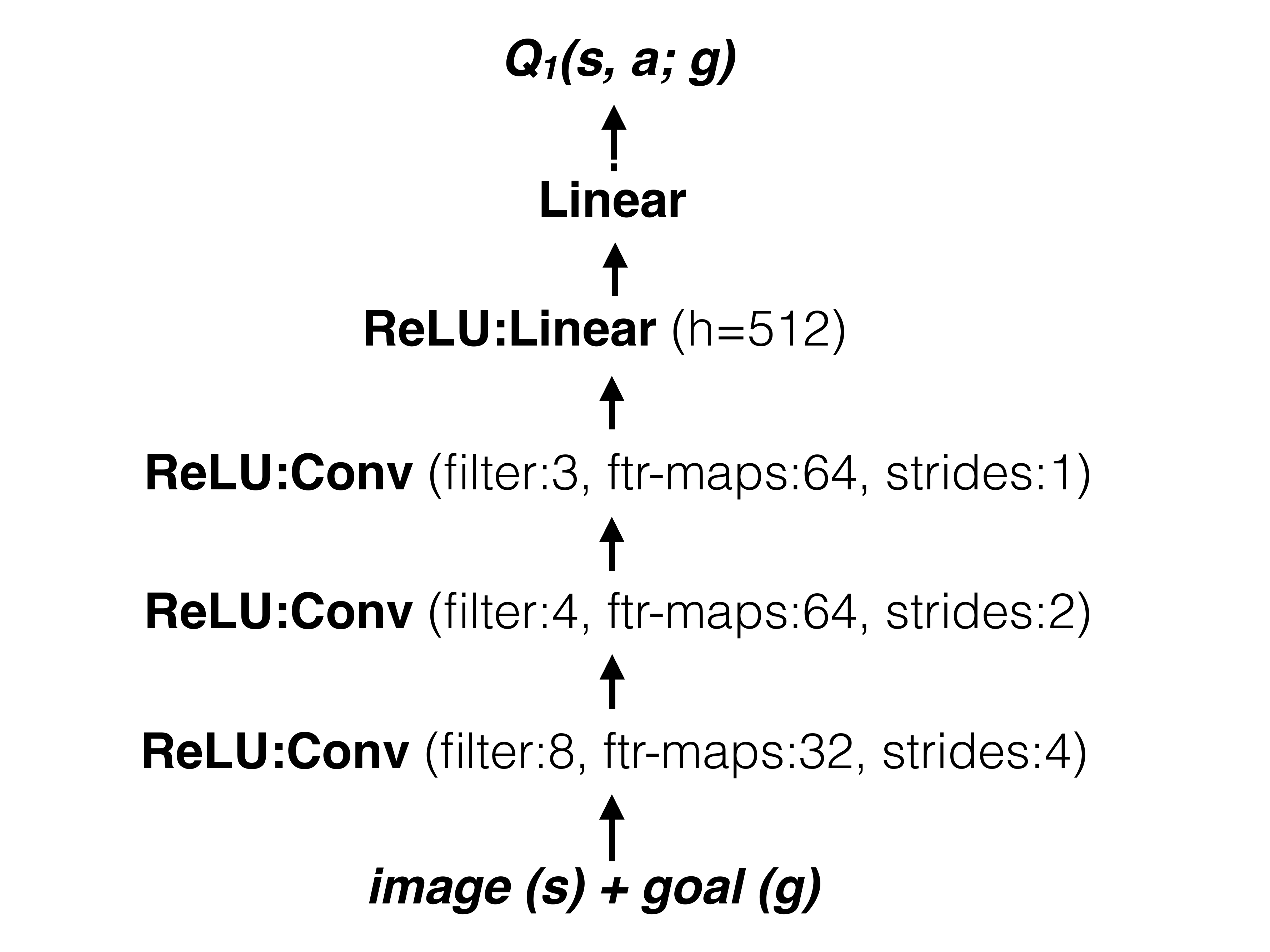}
	\caption{(a) A sample screen from the ATARI 2600 game called `Montezuma's Revenge'. (b) \textbf{Architecture}: DQN architecture for the controller ($Q_1$). A similar architecture produces $Q_2$ for the meta-controller (without goal as input). In practice, both these networks could share lower level features but we do not enforce this.}
\label{fig:montezuma_overview}
\end{figure*}

\paragraph{Game Description}
We consider `Montezuma's Revenge', an ATARI game with sparse, delayed rewards. The game (Figure \ref{fig:montezuma_overview}(a)) requires the player to navigate the explorer (in red) through several rooms while collecting treasures. In order to pass through doors (in the top right and top left corners of the figure), the player has to first pick up the key. The player has to then climb down the ladders on the right and move left towards the key, resulting in a long sequence of actions before receiving a reward (+100) for collecting the key. After this, navigating towards the door and opening it results in another reward (+300). 

% \begin{figure}[h]
%     \centering
%     \includegraphics[scale=0.39]{media/model.png}
%     \caption{Q1 and Q2 both have the same architecture except the input/output. Technically, both these network could share lower level features.}
%     \label{fig:syn_goals}
% \end{figure}

\begin{figure*}[h]
 \centering
        \centering
        \includegraphics[width=0.75\textwidth]{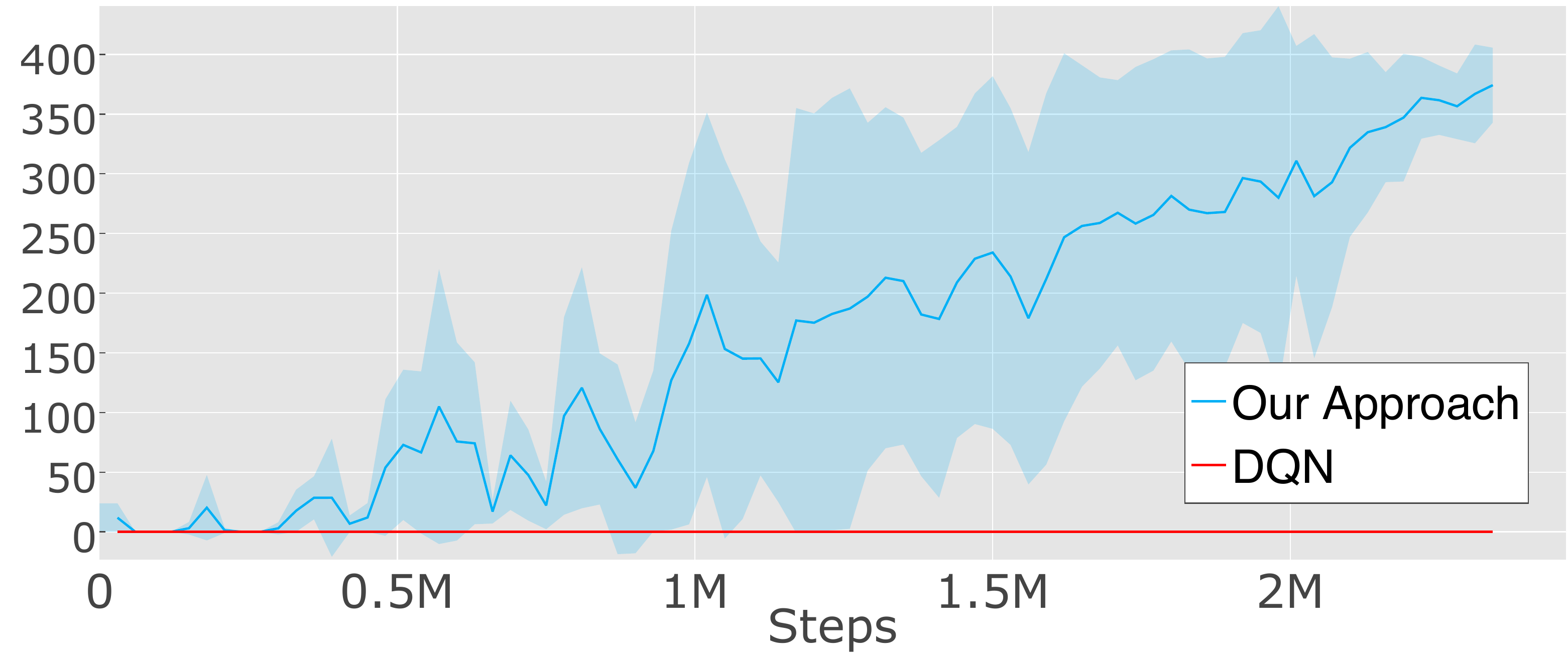}
        \caption*{(a) Total extrinsic reward}
        \label{fig:mz_reward}

        \centering
        \includegraphics[width=0.75\textwidth]{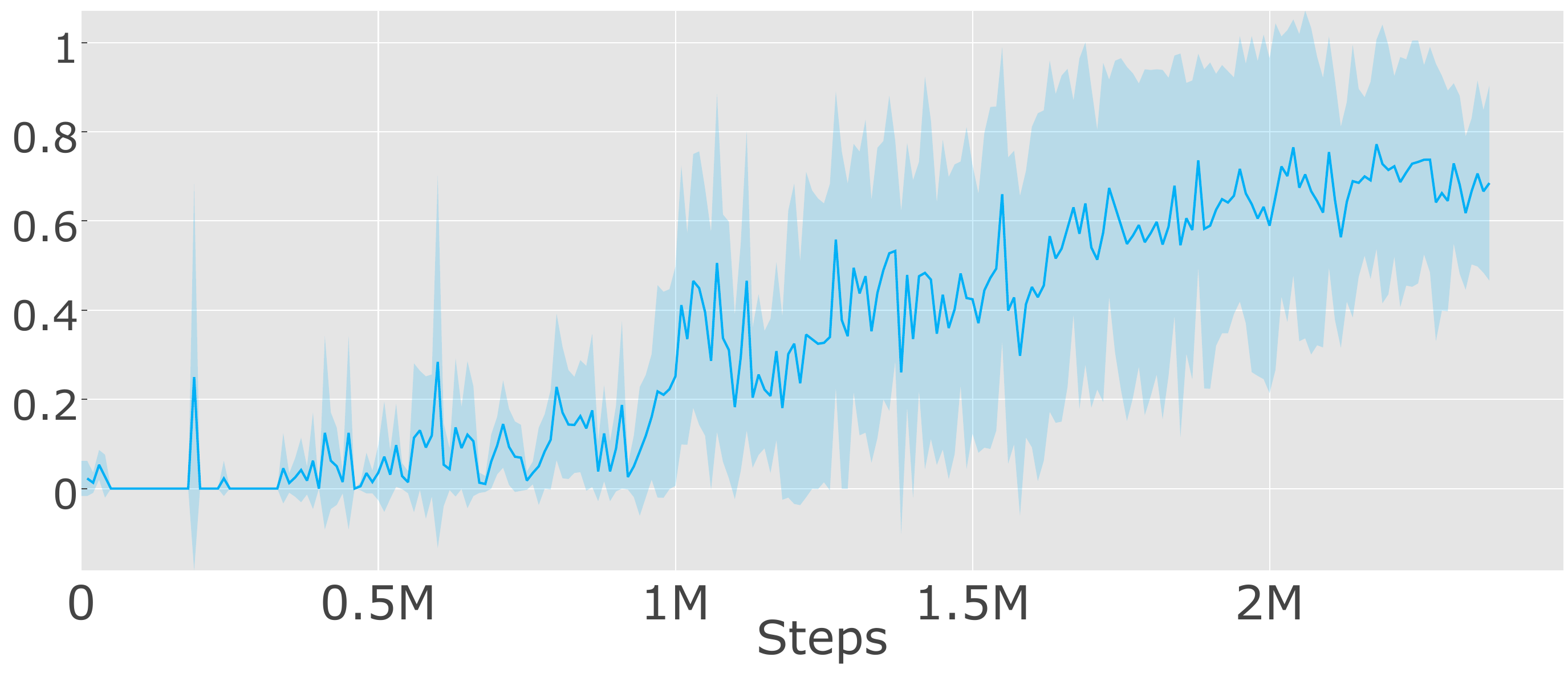}
        \caption*{(b) Success ratio for reaching the goal 'key'}
        \label{fig:mz_keysubgoal}

        \centering
        \includegraphics[width=0.75	\textwidth]{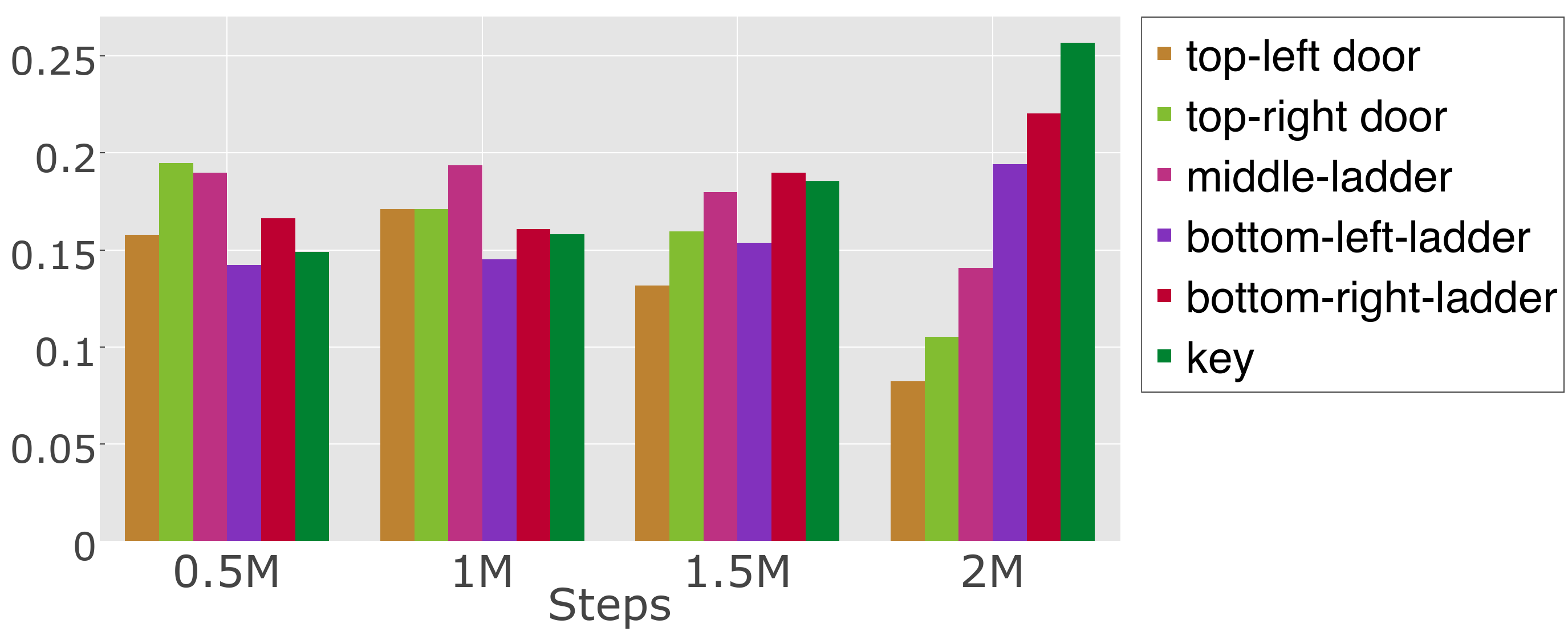}
        \caption*{(c) Success \% of different goals over time}
        \label{fig:mz_subgoalratio}

   \caption{\textbf{Results on Montezuma's Revenge:} These plots depict the joint training phase of the model. As described in Section~\ref{mzsec}, the first training phase pre-trains the lower level controller for about 2.3 million steps. The joint training learns to consistently get high rewards after additional 2 million steps as shown in \textbf{(a)}. \textbf{(b) Goal success ratio:} The agent learns to choose the key more often as training proceeds and is successful at achieving it. \textbf{(c) Goal statistics:} During early phases of joint training, all goals are equally preferred due to high exploration but as training proceeds, the agent learns to select appropriate goals such as the key and bottom-left door.}
    \label{fig:subgoal_mz}
\end{figure*}

Existing deep RL approaches fail to learn in this environment since the agent rarely reaches a state with non-zero reward. For instance, the basic DQN~\cite{mnih2015human} achieves a score of 0 while even the best performing system, Gorila DQN~\cite{nair2015massively}, manages only 4.16 on average. 

\paragraph{Setup} 
The agent needs intrinsic motivation to explore meaningful parts of the scene before it can learn about the advantage of getting the key for itself. Inspired by the developmental psychology literature \cite{spelke2007core} and object-oriented MDPs \cite{diuk2008object}, we use entities or objects in the scene to parameterize goals in this environment. Unsupervised detection of objects in visual scenes is an open problem in computer vision, although there has been recent progress in obtaining objects directly from image or motion data \cite{fragkiadaki2015learning,eslami2016attend,greff2015binding}. In this work, we built a custom object detector that provides plausible object candidates. The controller and meta-controller are convolutional neural networks (see Figure \ref{fig:montezuma_overview}(b)) that learn representations from raw pixel data. We use the Arcade Learning Environment~\cite{bellemare2012arcade} to perform experiments. 

The internal critic is defined in the space of $\langle entity_1, relation, entity_2 \rangle$, where $relation$ is a function over configurations of the entities. In our experiments, the agent is free to choose any $entity_2$. For instance, the agent is deemed to have completed a goal (and receives a reward) if the \text{agent} entity \textit{reaches} another entity such as the \textit{door}. Note that this notion of relational intrinsic rewards can be generalized to other settings. For instance, in the ATARI game `Asteroids', the agent could be rewarded when the bullet \textbf{reaches} the asteroid or if simply the ship never \textbf{reaches} an asteroid. In the game of `Pacman', the agent could be rewarded if the pellets on the screen are \textbf{reached}. In the most general case, we can potentially let the model evolve a parameterized intrinsic reward function given entities. We leave this for future work.

\paragraph{Model Architecture and Training}
As shown in Figure~\ref{fig:montezuma_overview}b, the model consists of stacked convolutional layers with rectified linear units (ReLU). The input to the meta-controller is a set of four consecutive images of size $84 \times 84$. To encode the goal output from the meta-controller, we append a binary mask of the goal location in image space along with the original 4 consecutive frames. This augmented input is passed to the controller. The experience replay memories $\mathcal{D}_1$ and $\mathcal{D}_2$ were set to be equal to $1\mathrm{E}{6}$ and $5\mathrm{E}{4}$ respectively. We set the learning rate to be $2.5\mathrm{E}{-4}$, with a discount rate of $0.99$. 
We follow a two phase training procedure -- (1) In the first phase, we set the exploration parameter $\epsilon_2$ of the meta-controller to 1 and train the controller on actions. This effectively leads to pre-training the controller so that it can learn to solve a subset of the goals. (2) In the second phase, we jointly train the controller and meta-controller.

\paragraph{Results}
Figure~\ref{fig:subgoal_mz}(a) shows reward progress from the joint training phase from which it is evident that the model starts gradually learning to both reach the key and open the door to get a reward of around +400 per episode. As shown in Figure~\ref{fig:subgoal_mz}(b), the agent learns to choose the key more often as training proceeds and is also successful at reaching it. As training proceeds, we observe that the agent first learns to perform the simpler goals (such as reaching the right door or the middle ladder) and then slowly starts learning the `harder' goals such as the key and the bottom ladders, which provide a path to higher rewards.
Figure~\ref{fig:subgoal_mz}(c) shows the evolution of the success rate of goals that are picked. At the end of training, we can see that the 'key', 'bottom-left-ladder' and 'bottom-right-ladders' are chosen increasingly more often. In order to scale-up to solve the entire game, several key ingredients are missing such as -- automatic discovery of objects from videos to aid goal parametrization we considered, a flexible short-term memory, ability to intermittently terminate ongoing options. 

\begin{figure*}[h]
    \centering
    \includegraphics[width=\textwidth]{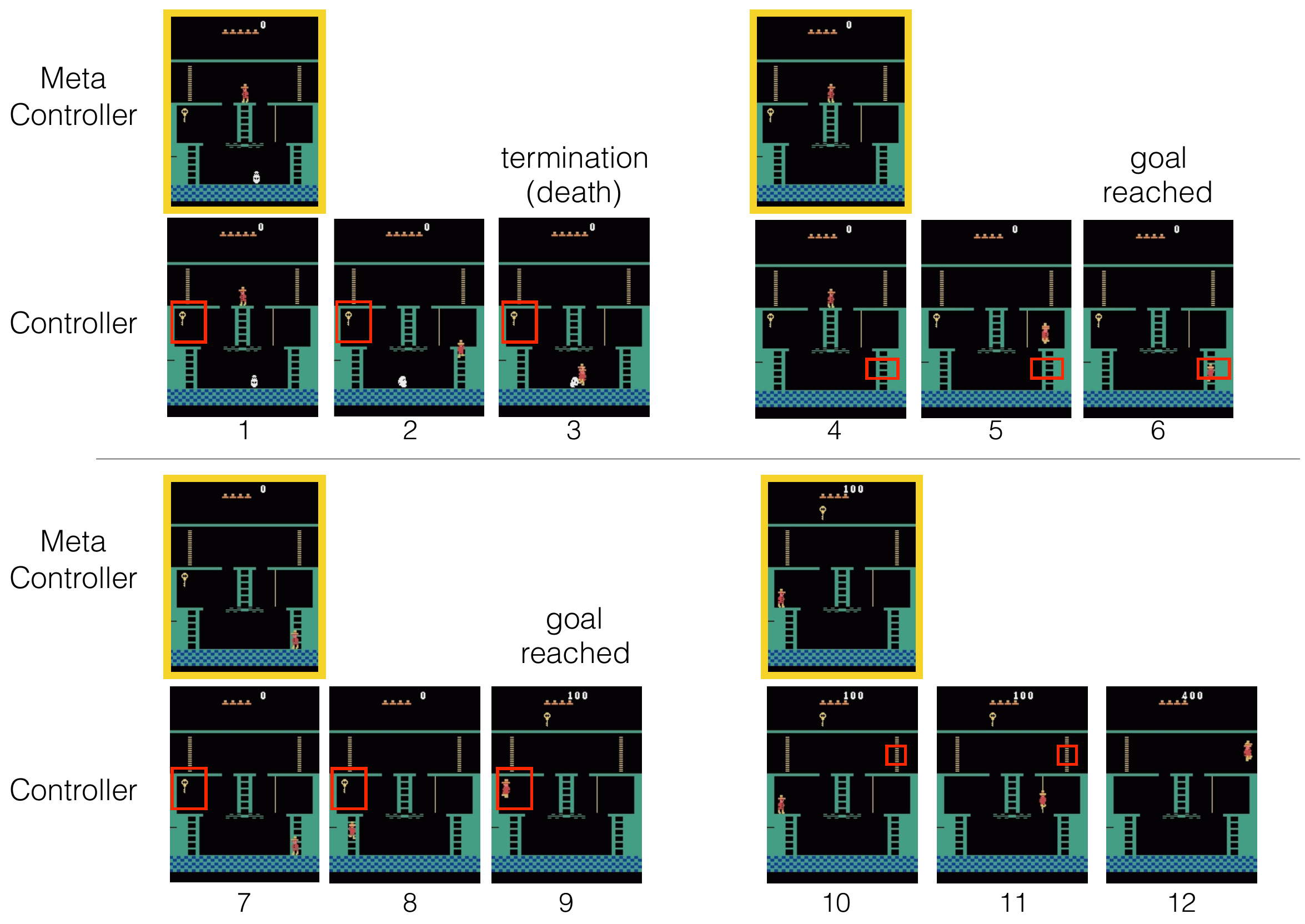}
    \caption{\textbf{Sample gameplay by our agent on Montezuma's Revenge:} The four quadrants are arranged in a temporally coherent manner (top-left, top-right, bottom-left and bottom-right). At the very beginning, the meta-controller chooses key as the goal (illustrated in \textit{red}). The controller then tries to satisfy this goal by taking a series of low level actions (only a subset shown) but fails due to colliding with the skull (the episode terminates here). The meta-controller then chooses the bottom-right ladder as the next goal and the controller terminates after reaching it. Subsequently, the meta-controller chooses the key and the top-right door and the controller is able to successfully achieve both these goals.}
    \label{fig:ills_mz}
\end{figure*}

We also show some screen-shots from a test run with our agent (with epsilon set to 0.1) in Figure~\ref{fig:ills_mz}, as well as a sample animation of the run.\footnote{Sample trajectory of a run on 'Montezuma's Revenge' --  \url{https://goo.gl/3Z64Ji} }

\section{Conclusion}

We have presented h-DQN, a framework consisting of hierarchical value functions operating at different time scales. Temporally decomposing the value function allows the agent to perform intrinsically motivated behavior, which in turn yields efficient exploration in environments with delayed rewards. We also observe that parameterizing intrinsic motivation in the space of entities and relations provides a promising avenue for building agents with temporally extended exploration. We also plan to explore alternative parameterizations of goals with h-DQN in the future.  

The current framework has several missing components including automatically disentangling objects from raw pixels and a short-term memory. The state abstractions learnt by vanilla deep-Q-networks are not structured or sufficiently compositional. There has been recent work \cite{eslami2016attend, greff2015binding, rezende2016one, kulkarni2015deep, whitney2016understanding, gregor2015draw, huang2015efficient} in using deep generative models to disentangle multiple factors of variations (objects, pose, location, etc) from pixel data. We hope that our work motivates the combination of deep generative models of images with h-DQN. Additionally, in order to handle longer range dependencies, the agent needs to store a history of previous goals, actions and representations. There has been some recent work in using recurrent networks in conjunction with reinforcement learning \cite{hausknecht2015deep, narasimhan2015language}. In order to scale-up our approach to harder non-Markovian settings, it will be necessary to incorporate a flexible episodic memory module.

% \begin{figure*}
%   \centering
%   (a) \includegraphics[width=4in]{media/overview.pdf}\\
%   (b) \includegraphics[width=4in]{media/NVNM.pdf}

%   \caption{TODO}.
%   \label{fig:overview}
% \end{figure*}

\section*{Acknowledgements}
We would like to thank Vaibhav Unhelkar, Ramya Ramakrishnan, Sam Gershman, Michael Littman, Vlad Firoiu, Will Whitney, Max Kleiman-Weiner and Pedro Tsividis for critical feedback and discussions. We are grateful to receive support from the Center for Brain, Machines and Minds (NSF STC award CCF - 1231216) and the MIT OpenMind team. 

% \bibliography{references}

\bibliographystyle{abbrv}
\small{\bibliography{references}}

\end{document}